\pdfoutput=1

\documentclass[11pt]{article}

\usepackage[]{emnlp2021}

\usepackage{times}
\usepackage{latexsym}
\usepackage{graphicx}

\usepackage[T1]{fontenc}

\usepackage[utf8]{inputenc}

\usepackage{microtype}

\title{YANMTT: Yet Another Neural Machine Translation Toolkit}

\author{
Raj Dabre\hspace{1em}
Eiichiro Sumita \\
National Institute of Information and Communications Technology, Kyoto, Japan\\
\texttt{\{raj.dabre, eiichiro.sumita\}@nict.go.jp}
}

\begin{document}
\maketitle
\begin{abstract}
In this paper we present our open-source neural machine translation (NMT) toolkit called ``Yet Another Neural Machine Translation Toolkit'' abbreviated as YANMTT\footnote{\url{https://github.com/prajdabre/yanmtt}} which is built on top of the Transformers library. Despite the growing importance of sequence to sequence pre-training there surprisingly few, if not none, well established toolkits that allow users to easily do pre-training. Toolkits such as Fairseq which do allow pre-training, have very large codebases and thus they are not beginner friendly. With regards to transfer learning via fine-tuning most toolkits do not explicitly allow the user to have control over what parts of the pre-trained models can be transferred. YANMTT aims to address these issues via the minimum amount of code to pre-train large scale NMT models, selectively transfer pre-trained parameters and fine-tune them, perform translation as well as extract representations and attentions for visualization and analyses. Apart from these core features our toolkit also provides other advanced functionalities such as but not limited to document/multi-source NMT, simultaneous NMT and model compression via distillation which we believe are relevant to the purpose behind our toolkit. 
\end{abstract}

\section{Introduction}
Neural machine translation (NMT) \cite{bahdanau15} is an end-to-end approach which is known to give state of the art results for a variety of language pairs. Thanks to the existence of publicly available toolkits such as OpenNMT, Fairseq, tensor2tensor, etc. NMT model training has become easier than ever. However, these toolkits have substantially grown in size over time and it is rather difficult for a beginner to dig into the code and understand the overall flow or make changes to the code base. Recently, the Transformers library by HuggingFace has become extremely popular and has made it possible for anyone to utilize publicly available models via scripts spanning a few lines. The most attractive feature is that pre-trained NMT models, MBART \cite{liu2020multilingual} being the most popular, can be easily downloaded and fine-tuned for a variety of applications which is a boon in low-resource settings. However, when it comes to pre-training models from scratch, one will be forced to write\footnote{At the time of creating this toolkit we noticed on the transformer library's github that several people requested a script for pre-training MBART models but none was available and the library developers did not express any strong enthusiasm to provide one.} their own code. 

While Fairseq allows one to pre-train MBART models, its codebase has grown extremely large which would make it difficult for NMT beginners to understand and potentially improve the flow of NMT pre-training. We feel that the lack of a simple script covering a pipeline for pre-training NMT models severely limits understanding and research into this important sub-field. Given the heavy adoption of the Transformers library we believe that it would good to have a toolkit built on top of it which enables pre-training, fine-tuning and related functionalities. We also note the following issues:

\noindent \textbf{1.} Publicly available MBART models are typically trained and shared by large companies such as Facebook, Google or HuggingFace but the increasing availability of data and computational resources makes it possible for researchers belonging to universities and smaller companies to pre-train their own models. We believe that pre-training own models is important because publicly available pre-trained models are often trained on generic corpora and should be adapted via additional pre-training so as to incorporate additional languages or for specializing them for particular domains. 

\noindent \textbf{2.} Distributed training, although easy to implement and extremely important during pre-training due to the sheer volume of training data, is heavily under-utilized due to lack of instructions. 

\noindent \textbf{3.} Pre-trained models, due to their large sizes, will slow down decoding and in order to be useful in low latency settings, they should be compressed via methods such as parameter tying and knowledge distillation. However there is a lack of a simple toolkit/script to explore this.

\noindent \textbf{4.} We noticed that existing toolkits simply begin with initialisation with pre-trained parameters followed by freezing certain components followed by fine-tuning. According to us, allowing users a greater degree of control over which components should and should not be transferred is important. 

\noindent \textbf{5.} Document/multi-source NMT is another way of leveraging additional information during training and is complementary to pre-training where additional information comes from monolingual documents. On the other hand simultaneous NMT helps in dealing with low latency settings and is complementary to model compression. Thus far we are not aware of toolkits that enable one to focus on the conjunction of these approaches with large scale pre-training.

In light of this situation we put together a toolkit, built on top of Transformers, consisting of simple scripts for distributed multilingual NMT pre-training, fine-tuning, parameter tying, compression, decoding and analysis. The NMT models themselves can be simultaneous or document/multi-source models. We call our toolkit \textbf{YANMTT} which stands for ``Yet Another Neural Machine Translation Toolkit''. Our toolkit is simple, minimal and sacrifices a bit of training efficiency for ease of understanding of the training pipeline. We have also heavily commented our code, often resorting to line by line comments, so that users know exactly what operation is being performed where. We hope that this beginner friendly nature of YANMTT should help entice more researchers into advancing the field of NMT pre-training.  

\section{Related Work}
Tensor2tensor \cite{vaswani-etal-2018-tensor2tensor} is a well developed library for training recurrent, convolutional as well as transformer models for a variety of sequence to sequence applications. However, it is no longer in active development due to the focus on Trax\footnote{\url{https://github.com/google/trax}}. While Tensor2tensor uses Tensorflow as a backend, Fairseq \cite{ott-etal-2019-Fairseq} is based on PyTorch \cite{NEURIPS2019_9015} and it allows one to train a variety of NMT models. Unlike Tensor2tensor, Fairseq contains all necessary functionality for pre-training NMT models but there is a severe lack of instructions for the same. Open-NMT \cite{klein-etal-2017-opennmt}, originally developed for recurrent NMT models is based on Tensorflow as well as PyTorch. THUMT \cite{zhang2017thumt} is a NMT training toolkit based on Tensorflow, PyTorch and Theano. Most recently, the Transformers \cite{wolf-etal-2020-transformers} library by Huggingface, based on PyTorch and Tensorflow has become extremely popular as it allows users to share trained models easily. In Transformers, the instructions for fine-tuning pre-trained models are abundant but to the best of our knowledge there is no complete script or  instruction for pre-training; a gap that we fill.

\section{Our Toolkit: YANMTT}
\begin{figure*}
    \centering
    \includegraphics[width=0.9\textwidth]{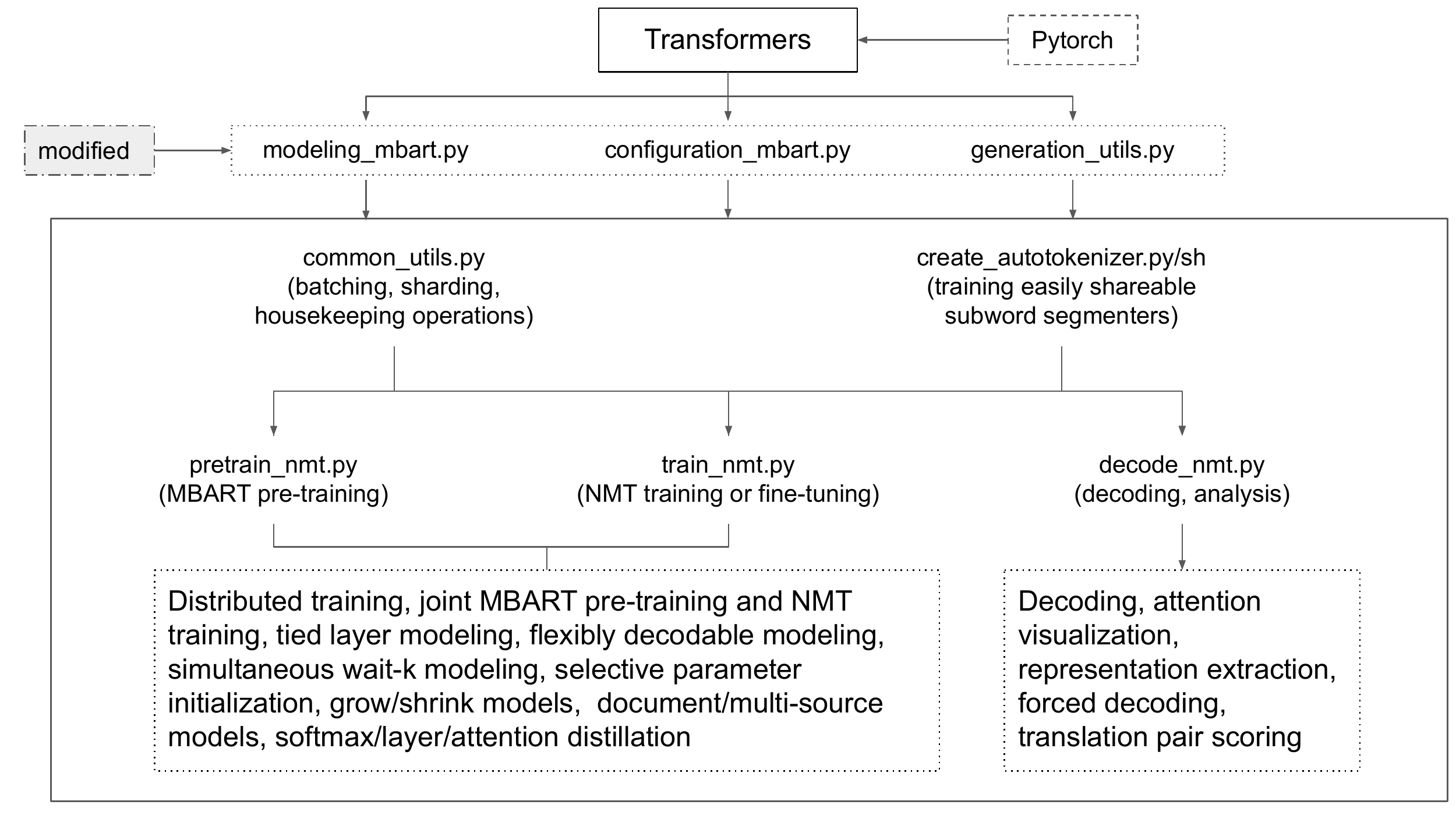}
    \caption{An overview of our YANMTT toolkit.}
    \label{fig:overview}
\end{figure*}

YANMTT, Yet Another Neural Machine Translation Toolkit, relies on the Transformers library v4.3.2 and uses PyTorch v1.7.1. We use only the MBART implementation (for now) from Transformers and write several wrapper scripts enabling multilingual NMT pre-training, controlled parameter transfer for fine-tuning, decoding, representation extraction and distillation. Apart from this we also modify the MBART modeling code to provide several relevant and advanced features. We provide the modified code (``transformers'' folder) along with our toolkit. We also provide example data and usage instructions in scripts (``examples'' folder) which will be made more comprehensive over time. We encourage the reader look at our toolkit\footnote{\url{https://github.com/prajdabre/yanmtt}} and watch the demo video\footnote{\url{https://youtu.be/UR9nkQz4pwc}} for a better understanding. Figure~\ref{fig:overview} contains an overview of our toolkit.

\subsection{Core Scripts and Features}
We describe the main scripts we wrote and augmentations to the MBART code in Transformers.

\subsection{create\_autotokenizer.sh/py}
These scripts govern the creation of a unigram SPM or BPE tokenizer using sentencepiece \cite{kudo-richardson-2018-sentencepiece}. This is wrapped around an AlbertTokenizer (for SPM) or MBartTokenizer (for BPE), special user defined tokens are added and a configuration file is created for use in the future via an AutoTokenizer. This tokenizer can be shared\footnote{\url{https://huggingface.co/transformers/model_sharing.html}} publicly.

\subsubsection{pretrain\_nmt.py}
This is used to train an MBART model using monolingual corpora and a pre-trained tokenizer. For MBART pre-training we provide users with the option to do text infilling as well as sentence permutation, both known to give the best results, however the latter is only performed when documents are available for training. We allow users to choose the amount of masking as well as lengths of word sequences to be infilled. This script can also allow one to load a MBART model trained by others and continue pre-training it on specific monolingual corpora. This should be useful when the model is required to be refined as more data and languages become available. We also enable joint MBART pre-training and regular NMT training. 

\subsubsection{train\_nmt.py}
This is used to either train a NMT model from scratch or fine-tune a pre-trained MBART or NMT model. Fine-tuning an NMT model can take the form of domain adaptation \cite{chu-etal-2017-empirical} or cross-lingual transfer \cite{DBLP:conf/emnlp/ZophYMK16:original,chu-dabre-2019}. It is possible to train unidirectional as well as multilingual multiway models \cite{johnson17,10.1145/3406095}. Different from most existing toolkits, we save separate model checkpoints for each translation direction being evaluated with the best development set BLEU score \cite{papineni-etal-2002-bleu} for that translation direction along with a single checkpoint with the highest average BLEU. We allow users to control the skew in corpora sizes via temperature based sampling \cite{arivazhagan2019massively}.

\subsection{decode\_nmt.py}
This is used to decode sentences using a trained model. We provide users with an option to mask spans of words in the input sentences when decoding using MBART models which can potentially variations of the original sentence. Additionally one can do translation pair scoring, forced decoding, encoder/decoder representation extraction and alignment visualization. Scoring should be helpful in corpora extraction where mined pairs can be ranked according to the model scores. On the other hand, the encoder/decoder representations may be analysed or visualized via appropriate toolkits\footnote{\url{https://projector.tensorflow.org}} or used for sentence classification.

\subsubsection{common\_utils.py}
This contains all housekeeping functions such as but not limited to data sharding, text infilling/masking, data batching, loss computation and attention visualization. Users are encouraged look into this script and modify the code as required.

\subsection{Advanced Features}
We now describe the advanced features of our toolkit which will enable users to train models at scale, train compressed models and have a greater degree of control over training.

\subsubsection{Training Models At Scale}
Using PyTorch's Distributed Data Processing wrapper, we enable training models on multiple gpus which may be distributed across $N$ machines and share a single file system. We assume head (index $0$) and children (indices $1$ to $N-1$) machines. Users need only note the IP address of the head and first launch the training command on the it specifying the index ($0$), number of machines ($N$) and gpus per machine. Thereafter the command should be launched on children specifying indices ($1$ to $N-1$), number of machines ($N$), gpus per machine and the IP address of the head. PyTorch allows for automatic mixed precision which we also incorporate in our toolkit and found that it gives modest speedups during training\footnote{As automatic mixed precision is different from pure 16-bit training the speedups can be as much as 40\%. Furthermore, we observed that training behaves strangely when default hyperparameters are used when training on a large number of GPUs so the user should be careful and default to 32-bit training in the worst case.}. When training in a multi-gpu or a distributed fashion users may use our automatic simple data sharing mechanism that ensures that each GPU automatically has access to a subset of the entire data. We have ensured that using the distributed training is as simple as possible and have also annotated the parts of our code so that users may learn and implement it in their own pipelines.

\subsubsection{Controlled Parameter Initialization}
Typically all pre-trained model parameters are used for fine-tuning which means that the final model will be as large as the pre-trained model. Furthermore, not all pre-trained parameters might be needed for the fine-tuning task, especially when the pre-training and fine-tuning vocabulary are different. To this end we have implemented functionalities which allow users to specify which components should be initialized using pre-trained parameters and which should be randomly initialized. This allows the refined model can have more or fewer layers than the existing MBART model.  It is also possible to initialize the embedding layer for subwords that are common between the current and previous tokenizer while initializing the remaining embeddings randomly. We believe that this should allow users to start out with a pre-trained model with a particular number of layers and vocabulary sizes and then grow or shrink them when doing fine-tuning on other data. It makes sense to expand the number of layers and vocabulary sizes when additional data becomes available. In such a case, the new layers may be initialized with existing parameters of existing layers or with random parameters. On other hand, shrinking the number of layers makes sense when we want to sacrifice some performance but have faster models. Naturally, faster models can be obtained via sequence distillation which we have also implemented.  

\subsubsection{Training Compact Models}
While initializing smaller models with a subset of the parameters of a larger model may help in obtaining compact models with some loss in performance, we alternatively enable users to train highly compact models from scratch by implementing recurrently stacked layers \cite{Dabre_Fujita_2019}. This involves tying parameters of layers and the resultant models tend to be 50-70\% smaller depending on the tying configuration, number of layers and hidden sizes. This concept of tying parameters across layers has also been shown to work well for pre-trained AlBERT models in text classification tasks \cite{lan2019albert}. On a related note, we have also implemented the training time compute intensive multi-layer softmaxing \cite{dabre-etal-2020-balancing} approach where cross entropy loss is computed using each decoder layer's output. This means that fewer decoder layers can be used whenever faster translations are desired.

Apart from parameter tying we also allow users to perform knowledge distillation \cite{HinVin15Distilling,kim-rush-2016-sequence} in 3 different ways. The first way is the classic, teacher-student cross-entropy minimization \cite{HinVin15Distilling} where the teacher's (larger model) probability distribution is used in place of the labels during training the child (smaller model). The second way involves minimizing the mean squared difference between the hidden layer representations of the teachers and students \cite{sun-etal-2019-patient}. The third and final way involves minimizing the cross entropy between the self/cross attentions of the teacher and students \cite{tian2019crd}. For the second and third way, one can specify the mappings between the teacher and student layers. Furthermore, it is possible to use a combination of the 3 ways. Ultimately we expect that combining layer tying and distillation should help users train highly compact models.

\subsection{Advanced NMT Models}
We also make some changes to the MBART code itself so as to allow users to train simultaneous, document/multi-source translation models. We believe that these aspects are related to the overall theme of this toolkit for the following reasons: Since pre-trained models tend to be extremely large and thereby compute intensive we believe that when fine-tuning them, it should be important to consider simultaneous translation. Just like pre-trained models allow leveraging information from additional sources (monolingual data), leveraging information from the translation context should also be focused on. Indeed, models like MBART are typically pre-trained on documents and so they should be very helpful for document NMT and thus our document NMT implementation should help.

\subsubsection{wait-k Simultaneous NMT}
While compact models are useful in a low latency translation setting, simultaneous NMT models in production settings are the most attractive options. Given the focus on speeding up NMT models we made slight changes to the modeling code so as to allow users to simulate wait-k simultaneous NMT models \cite{ma-etal-2019-stacl} where decoding begins when $k$ encoder tokens are available. Users can train wait-k models with fixed as well as randomly sampled values of $k$ or train full sentence models with unidirectional\footnote{This is simply implemented via a unidirectional self attention mask in the encoder.} encoders and decode them with any desired value of $k$. We hope that this will enable and encourage researchers to focus on ideas that combine model compression and simultaneous translation and push the boundaries of high quality and low latency NMT.

\subsubsection{Document/Multi-Source NMT}
To train document level models, users should provide the sentence to be translated and its context. The source sentence and context are encoded and then combined. We have implemented a variety of combination approaches of which two of the well known ones are "multi-source combination" and "context attention combination". In multi-source combination, the document context is treated treated as an additional source sentence and their representations are combined via a weighted sum of the attentions computed on them by the decoder as proposed by \citet{N16-1004}. In context attention combination \cite{voita-etal-2018-context}, self attention is computed on the sentence representation and cross attention is computed between the sentence (query) and the context (key/value) representations. Thereafter, these attentions are combined via the aforementioned weighting mechanism. The major difference is that in the former method the representations are combined in the decoder and in the latter method the representations are combined in the encoder. Our document NMT implementations can be directly used for multi-source NMT \cite{Dabre-MTS2017,N16-1004} by replacing the document context with sentences corresponding to the additional source language. Furthermore, by passing flags corresponding to simultaneous NMT to the training script, one can train multi-source simultaneous wait-k NMT \cite{dabre2021simultaneous} or simultaneous document NMT. Ultimately all these models can be compressed via distillation facilities we provide and we believe that this falls within our goals towards providing users to develop models that will be compact and used in real time translation settings. 

\section{Experiments}
We now describe basic experiments we conducted to test the working of the core functionalities of our toolkit. Due to lack of space we do not show results for the advanced features.

\subsection{Datasets and Pre-processing}
We experiment with Workshop on Asian Translation (WAT) \cite{nakazawa2018overview} datasets such as ASPEC Japanese--English dataset (~3M pairs), IIT Bombay Hindi--English dataset (~1.5M pairs) and WAT 2021's MultiIndicMT\footnote{\url{http://lotus.kuee.kyoto-u.ac.jp/WAT/indic-multilingual/}} dataset (10 Indian languages to English; ~23k to 50k pairs depending on the language pair from the PMI subset of the whole dataset). We consider the ASPEC and IIT Bombay datasets as resource rich and the MultiIndicMT dataset as resource poor. For MBART pre-training we experiment with the IndicCorp monolingual dataset covering 11 Indic languages and English (~1M to 63M sentences depending on the language; ~450M sentences in total). We segment Japanese sentences using Mecab. Except this we do not do any other pre-processing. We use shared subword vocabularies obtained using sentencepiece. The vocabulary sizes are as follows: 32,000 for Japanese--English,  8,000 for Hindi--English and 64,000 for all languages in MultiIndicMT. The vocabulary for the MultiIndicMT task is obtained from the IndicCorp dataset because we train our MBART model on IndicCorp and fine-tune it on MultiIndicMT data. 

\subsection{Training Details}
We use the base (6 layers, hidden-filter sizes of 512-2048) architecture when pre-training is not used. For the MBART model and fine-tuning we use the big architecture (6 layers, hidden-filter sizes of 1024-4096). We use the MBART model for improving the performance of the MultiIndicMT task and it is trained using text infilling using most settings similar to \citet{liu2020multilingual}. The Japanese--English and Hindi--English models are trained on 8 V-100 GPUs on one machine, the MBART model is trained on 48 V-100 GPUs spread over 6 machines and the Indic--English models (baseline and fine-tuned) is trained on 1 V-100 GPU. For pre-training we use batch sizes of 4096 tokens whereas for regular MT training and fine-tuning we use batch sizes of 2048 tokens. We use learning rates of 0.001 with 16,000 warmup steps. We train our translation models till convergence on the development set BLEU score \cite{papineni-etal-2002-bleu} which is computed every 1,000 batches and choose the model checkpoint corresponding to the best development set score for decoding. On the other hand we train our MBART model for roughly 1 epoch\footnote{We use a data sampling temperature of 5 to ensure that the model does not focus on the languages with abundant parallel corpora. However this means that 1 epoch involves more than the approximately 450M monolingual sentences.} corresponding to 140,000 batches per GPU which need roughly 1 day to finish. For decoding we use beam-search with beam of size 4 and length penalty of 1.0. Unless mentioned otherwise we use our script's default hyperparameters which we have tried to keep as similar as possible to default ones in previous works. All BLEU scores in the toolkit are computed using SacreBleu \cite{post-2018-call} but the scores we report in this paper obtained using WAT's evaluation interface\footnote{\url{http://lotus.kuee.kyoto-u.ac.jp/WAT/evaluation/index.html}} for ease of comparison other people's submissions.

\subsection{Results}
\subsubsection{Resource Rich Tasks}
For Japanese to English and English to Japanese translation on the ASPEC dataset we obtained BLEU scores of 28.35 and 39.96. For Hindi to English and English to Hindi translation on the IITB dataset we obtained BLEU scores of 19.26 and 18.94. Comparing against transformer model scores obtained via other toolkits on the leaderboard shows that the scores we obtain are comparable.

\begin{table}[]
    \centering
    \begin{tabular}{c|c|c}
        \textbf{Direction} &  \textbf{Baseline} & \textbf{Fine-tuned}\\\hline\hline
        hi$\rightarrow$ en &  28.21 & \textbf{35.80}\\
        mr$\rightarrow$ en & 15.10 & \textbf{25.45}\\
        bn$\rightarrow$ en & 11.27 & \textbf{21.37}\\
        kn$\rightarrow$ en & 20.33 & \textbf{29.29}\\\hline\hline
        en$\rightarrow$ hi & 23.31 & \textbf{29.59}\\
        en$\rightarrow$ mr & 8.82 & \textbf{14.69}\\
        en$\rightarrow$ bn & 5.58 & \textbf{10.59}\\
        en$\rightarrow$ kn & 10.11 & \textbf{16.13}\\\hline\hline
    \end{tabular}
    \caption{Results of baseline models and models trained via fine-tuning on MBART.}
    \label{tab:indicresults}
\end{table}
\subsubsection{Resource Poor Tasks}
Although our MBART model was trained on 11 Indic languages and English, the MultiIndicMT task contained parallel corpora for 10 Indic languages and English. Due to lack of space we report results for Hindi--English (hi--en), Marathi--English (mr--en), Bengali--English (bn--en) and Kannada--English (kn--en). The results for baseline and fine-tuned models are shown in Table~\ref{tab:indicresults}. It is clear that fine-tuning helps in low-resource settings.


\section{Conclusion and Future Work}
We have presented our open-source toolkit called "Yet Another Neural Machine Translation Toolkit", also known as YANMTT. YANMTT allows users to pre-train and fine-tune their own multilingual sequence to sequence models by leveraging a large number of GPUs. We also implemented functionalities for compressing large models via selective parameter transfer and knowledge distillation approaches. Additionally we have provided functionalities for simultaneous and document/multi-source NMT. In the future we plan to support additional NMT models which can work with long sequences. We also plan to implement latest simultaneous NMT and document NMT approaches. We also plan to improve the logging facilities so that users may be able to access gradients and other model internal information during training.

\bibliography{anthology,custom}
\bibliographystyle{acl_natbib}

\end{document}